\documentclass[sigconf]{acmart}
\renewcommand\footnotetextcopyrightpermission[1]{} 

\usepackage{textcomp}
\usepackage{xcolor}
\usepackage{graphicx}
\usepackage{graphics}
\usepackage{subcaption}
\usepackage{balance}
\usepackage[ruled,linesnumbered]{algorithm2e}
\usepackage{multirow}
\usepackage{makecell}
\usepackage{float}
\usepackage{hyperref}

\usepackage{subcaption}
\begin{document}
\pagestyle{plain} 
\title{Skip the Benchmark: Generating System-Level High-Level Synthesis Data using Generative Machine Learning}

\author{Yuchao Liao}
\affiliation{
  \institution{Electrical and Computer Engineering\\University of Arizona}
  \city{Tucson}
  \country{Arizona, USA}}
\email{yuchaoliao@arizona.edu}

\author{Tosiron Adegbija}
\affiliation{
  \institution{Electrical and Computer Engineering\\University of Arizona}
  \city{Tucson}
  \country{Arizona, USA}}
\email{tosiron@arizona.edu}

\author{Roman Lysecky}
\affiliation{
  \institution{Electrical and Computer Engineering\\University of Arizona}
  \city{Tucson}
  \country{Arizona, USA}}
\email{rlysecky@arizona.edu}

\author{Ravi Tandon}
\affiliation{
  \institution{Electrical and Computer Engineering\\University of Arizona}
  \city{Tucson}
  \country{Arizona, USA}}
\email{tandonr@arizona.edu}

\renewcommand{\shortauthors}{Liao et al.}


\begin{abstract}
High-Level Synthesis (HLS) Design Space Exploration (DSE) is a widely accepted approach for efficiently exploring Pareto-optimal and optimal hardware solutions during the HLS process. Several HLS benchmarks and datasets are available for the research community to evaluate their methodologies. Unfortunately, these resources are limited and may not be sufficient for complex, multi-component system-level explorations. Generating new data using existing HLS benchmarks can be cumbersome, given the expertise and time required to effectively generate data for different HLS designs and directives. As a result, synthetic data has been used in prior work to evaluate system-level HLS DSE. However, the fidelity of the synthetic data to real data is often unclear, leading to uncertainty about the quality of system-level HLS DSE. This paper proposes a novel approach, called \textit{Vaegan}, that employs generative machine learning to generate synthetic data that is robust enough to support complex system-level HLS DSE experiments that would be unattainable with only the currently available data. We explore and adapt a Variational Autoencoder (VAE) and Generative Adversarial Network (GAN) for this task and evaluate our approach using state-of-the-art datasets and metrics. We compare our approach to prior works and show that Vaegan effectively generates synthetic HLS data that closely mirrors the ground truth's distribution.
\end{abstract}


\maketitle

\date{July 2023}

\begin{figure}[t]
\centering
\includegraphics[width=0.9\columnwidth,keepaspectratio]{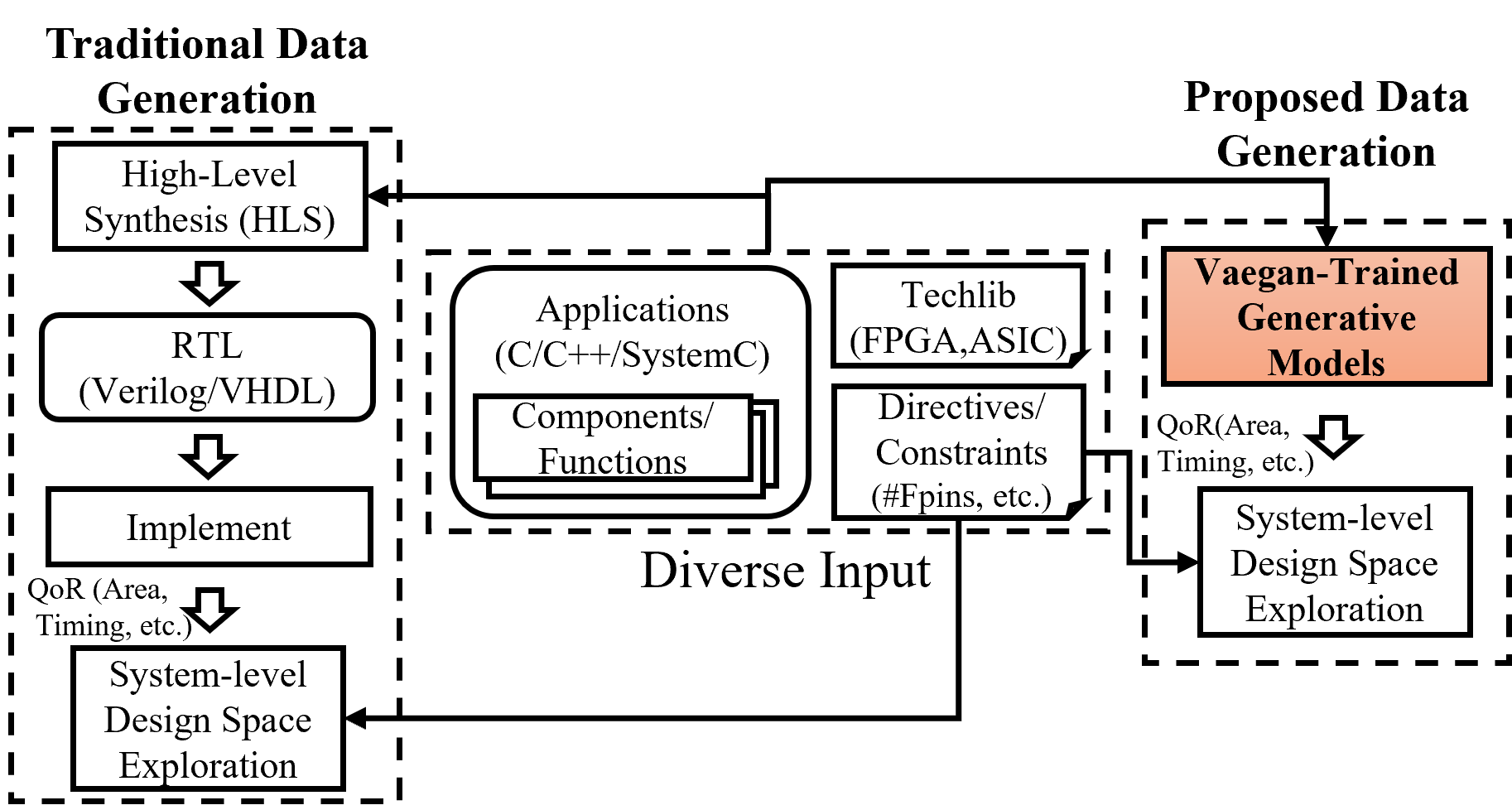}
\caption{Proposed Vaegan approach to generating synthetic data for system-level HLS design space exploration compared to the traditional approach}
\label{fig:Vaegan_use}
\end{figure}

\begin{figure*}[t]
\centering
\includegraphics[width=0.8\textwidth,keepaspectratio]{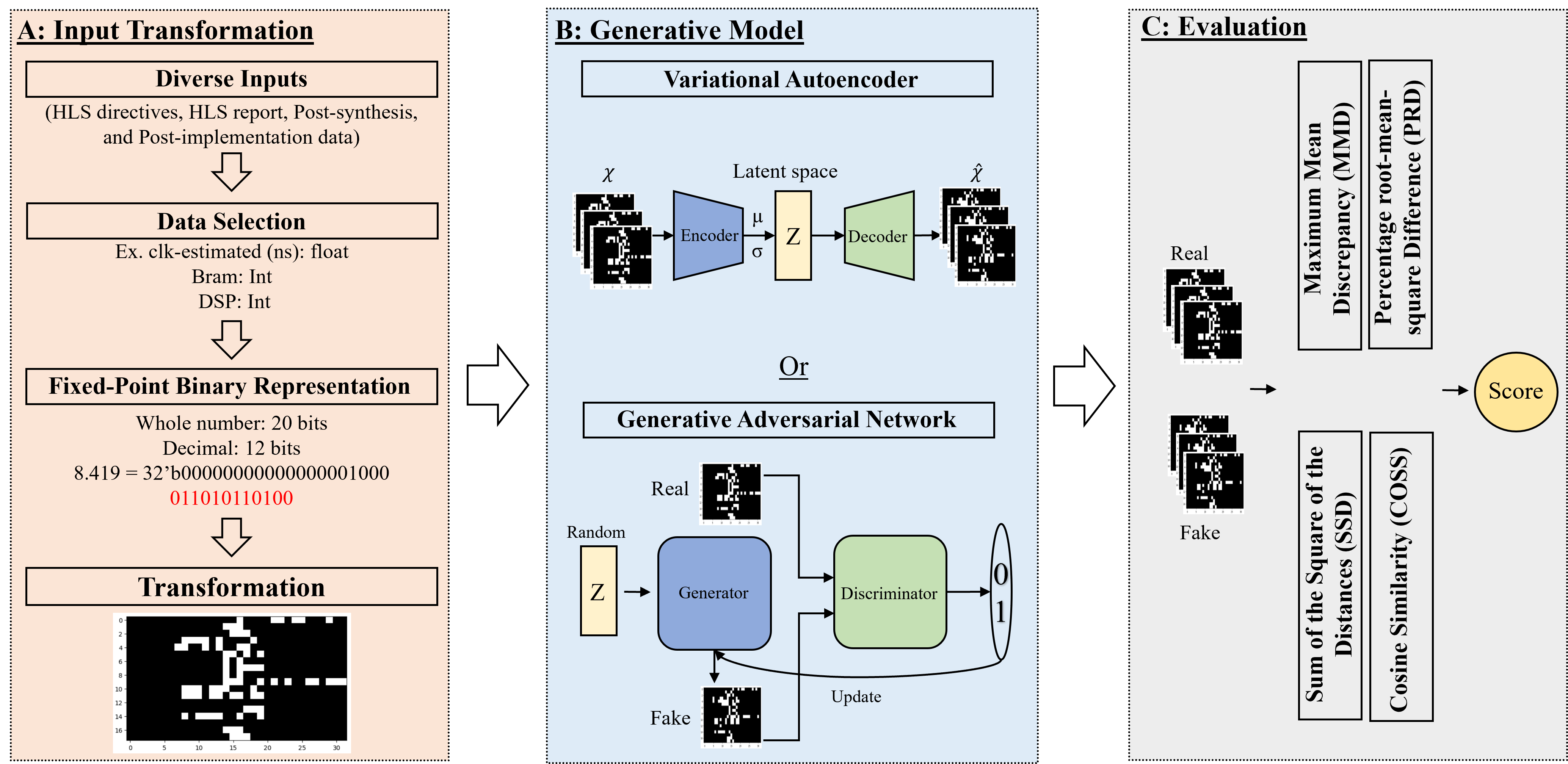}
\caption{Vaegan comprises three stages: (A) formatting and transforming diverse input data (HLS directives, HLS report estimation, post-synthesis, and post-implementation data) into a format readable by the network. (B) employing ML (here, a Variational Autoencoder (VAE) or a Generative Adversarial Network (GAN)) to generate synthetic HLS data. (C) evaluating the generated synthetic HLS data.}
\label{fig:methodology}
\end{figure*}

\section{Introduction}\label{sec:intro}
High-level synthesis (HLS) is a popular approach to designing, synthesizing, and optimizing hardware systems. HLS is often used to design embedded systems, such as medical devices, autonomous vehicles, and, more generally, the Internet of Things (IoT). Using existing HLS tools (like Vitis HLS), designers can develop application-specific embedded systems using high-level languages (e.g., C/C++) and map them to hardware register-transfer level (RTL) languages (e.g., Verilog, VHDL), thereby improving design productivity and reducing the design time/cost \cite{HLSsurveySchafer2022,LiaoDSE2023}. HLS tools allow designers to select different directives such as loop unrolling factors, memory binding, function inline, target frequency, etc. Each modification of directives creates a different design configuration, leading to a larger design space.

HLS design space exploration (DSE) \cite{HLSsurveySchafer2022} aims to identify the Pareto-optimal or optimal design solutions, considering factors such as performance, area, and power at both the system and component levels. State-of-the-art HLS DSE approaches use machine learning (ML) or heuristic-based methods to identify component-level \cite{wu2022ironman,GOSWAMI2023116} or system-level (multi-component) \cite{LiaoDSE2023} Pareto-optimal configurations. Training ML models and evaluating each heuristic requires extensive data. Several HLS benchmarks and datasets are available to the community for evaluating these methodologies \cite{Zhou2018Rosetta,pouchet2012polybench,Reagen2014MachSuite,wei2023hlsdataset,Yuko2008CHStone,ferretti2021db4hls}. However, these existing benchmarks and datasets focus on component/function-level computations without including system characteristics (e.g., end-to-end timing constraints) that may be required in real-world HLS usage scenarios. As such, these datasets cannot meet all experimental conditions, particularly for complex real-world embedded systems that require meeting timing constraints and minimizing energy consumption. For instance, as illustrated in Fig. \ref{fig:Vaegan_use}, performing a system-level HLS DSE analysis of a multi-component system (e.g., a wearable pregnancy monitoring system or an autonomous braking system) requires extensive synthesis, implementation, and validation of each component. These systems, initially generated using an HLS tool (e.g., Vitis HLS), are synthesized and implemented with an RTL tool (e.g., Xilinx Vivado) targeting a specific Field-Programmable Gate Array (FPGA). Collecting data for each system configuration is often prohibitively time-consuming.

Despite researchers dedicating considerable time to data collection, datasets targeting only a few FPGA boards and a specific HLS tool may prove insufficient for developing accurate ML models or heuristics for DSE. These models and heuristics are crucial for predicting hardware implementation results from HLS directives (e.g., loop unrolling factor, loop pipeline, array partition) or evaluating heuristic algorithms across diverse HLS tools and FPGA boards. Given the complexity of these unexplored conditions (e.g., embedded systems' time constraints), previous works \cite{LiaoDSE2023,wu2022ironman} have demonstrated that synthetic data can be effective in expanding the design space and evaluating HLS DSE methodologies.

Synthetic data offers numerous benefits for both ML-based and heuristic-based HLS DSE. For instance, it is cost effective and time efficient, and provides data diversity. Using both synthetic and real data for training can enhance the robustness of ML models, among other advantages \cite{lu2023machine}. However, creating synthetic HLS data and achieving \textit{high fidelity} (i.e., a similar data distribution between real and synthetic data) is challenging. Current works that use synthetic data for HLS DSE do not quantify the fidelity of the synthetic data they use, thus calling into question the effectiveness of the DSE evaluation. Existing approaches for generating synthetic data may result in low-fidelity data. We address this critical challenge using an innovative method for generating high-fidelity synthetic HLS data to support HLS DSE research.

In this paper, we propose a novel approach, called \textit{Vaegan}\footnote{The code is available at \url{https://github.com/yuchaoliao/VAEGAN.git}.}, to simplify the generation of synthetic data for system-level HLS DSE (Fig. \ref{fig:Vaegan_use}). Vaegan initially formulates and transforms diverse real HLS data, sourced from HLS directives, HLS report estimation, post-synthesis, and post-implementation data into a binary input format. The approach then employs generative machine learning---we explored Variational Autoencoder (VAE) \cite{kingma2022autoencoding} and Generative Adversarial Network (GAN) \cite{goodfellow2020GAN})---to generate and analyze the synthetic data. Vaegan allows designers to use and transform their own ground truth data to generate synthetic data. We evaluate our work using widely recognized metrics, and compare it both qualitatively and quantitatively to real HLS data and two prior works involving synthetic data \cite{LiaoDSE2023,wharrie2022hapnest}. Experimental results show that, compared to prior work, Vaegan effectively generates high-fidelity synthetic data that improves the Maximum Mean Discrepancy (MMD) score from real HLS data by 44.05\%. We also demonstrate the practical usefulness of the Vaegan approach through a case study that involves generating complex, synthetic HLS data for a wearable pregnancy monitoring system. Our results indicate that this synthetic data expands the Pareto frontier, uncovering new Pareto-optimal solutions that were not present in the original dataset, and extending both the quality and the quantity of the original dataset for design space exploration.

\section{Related Work}\label{sec:relatedwork}
State-of-the-art high-level synthesis (HLS) benchmarks feature many low- and top-level kernels. Examples include Rosetta \cite{Zhou2018Rosetta}, MachSuite \cite{Reagen2014MachSuite}, and Polybench \cite{pouchet2012polybench}. While recent HLS datasets, such as db4hls \cite{ferretti2021db4hls} and HLSDataset \cite{wei2023hlsdataset}, are highly valuable, they partially implement the above benchmarks and are limited in their diversity of experimental scenarios. For instance, HLSDataset omits data for execution cycles and post-implementation critical paths which are essential for energy calculation, while db4hls lacks post-synthesis and post-implementation results. Both datasets concentrate on a limited set of workloads and devices, with none considering real-world multi-component embedded systems. Consequently, HLS DSE research has turned to using synthetic data (in addition to real data) for evaluation \cite{LiaoDSE2023,wu2022ironman,liao2022high}. For example, Liao et al.  \cite{LiaoDSE2023} randomly generated post-implementation data based on real-world systems for system-level HLS DSE and Wu et al. \cite{wu2022ironman} randomly generated data flow graphs (DFGs) during early-stage HLS exploration. However, these works do not analyze the fidelity of the generated synthetic data to real HLS data. Our examination of the approaches employed in these works reveals a critical discrepancy between the synthetic and real HLS data, highlighting the need for a novel approach to generate and assess synthetic HLS data.

There are many related works on leveraging generative ML models for generating synthetic data in fields such as Natural Language Processing (NLP), vision, healthcare, voice, etc. \cite{lu2023machine}. However, none of them pertain to the generation of HLS data. Variational Autoencoder (VAE) \cite{kingma2022autoencoding} and Generative Adversarial Network (GAN) \cite{goodfellow2020GAN} are the most common ML frameworks used for synthetic data generation. VAE is a type of generative model that uses a probabilistic approach to compress data into a lower-dimensional space (encoding) and then generates new data by decoding from this latent space. GAN is composed of two neural networks---a generator and a discriminator---competing against each other, where the generator attempts to produce synthetic data that can deceive the discriminator network into believing it is real. Both models are explained in Section \ref{Sec:Generative models} along with how we tailor them for HLS data.To our knowledge, ours is the first work that tackles the challenge of generating high-fidelity system-level synthetic data for HLS DSE.

\section{Methodology}
Fig. \ref{fig:methodology} illustrates the overall design flow of Vaegan, which consists of three stages. The first stage involves formalizing the input and output. Diverse input data, such as clock, area, and power are transformed into a format readable by the network. The second stage employs a state-of-the-art generative network, specifically a Variational Autoencoder (VAE) or a Generative Adversarial Network (GAN), fine-tuned to generate the desired synthetic HLS data. The final stage evaluates the models and the generated synthetic data using state-of-the-art metrics. Here, we elaborate on these stages.

\subsection{Input Transformation}\label{Sec:inputformat}
Fig. \ref{fig:methodology}(A) shows the flow of input transformation. Vaegan begins by constructing the appropriate input sets, a task complicated by the intricate HLS process and the variety of data types. Our approach aims to enable designers to generate their own synthetic HLS data. As such, the first challenge in this step involves selecting critical variables for data generation. A typical HLS design point might comprise C/C++ source code, synthesis directives, Intermediate Representation (IR) of the source code, HLS report estimation data, post-synthesis data, and post-implementation data. Depending on their specific focus, designers may select any subset of these data. For instance, ML-based HLS DSE approaches often predict area and performance from the synthesis directives and IR to either post-synthesis or post-implementation data \cite{wu2022ironman,GOSWAMI2023116}. To evaluate Vaegan, we use HLS directives, HLS report estimation, post-synthesis, and post-implementation data.

HLS synthesis directives usually consist of loop unrolling factors, loop pipelining, array partitioning, memory type, function inlining, target frequency, etc. HLS report estimation data includes estimated synthesis results. These estimated results resemble actual implementation results from tools like Xilinx Vivado. Post-synthesis and post-implementation data, derived from synthesis tool reports, include metrics for execution cycles, critical path, area, power, and target frequency. Designer-selected data are combined into a single model input. Given the limitations of existing benchmarks and datasets, designers may need to manually generate real data as a ground truth for the ML model.

The second challenge is transforming diverse input data, such as resource usage (integers), power (floating-point numbers), and HLS directives (options with integers), into a network-readable format. Firstly, We propose converting each input variable to a binary fixed-point format, with the precision determined by the real data. In our experiments (Section \ref{sec:experiments}), we used a 32-bit representation, with 20 and 12 bits for the integer and fractional portions, respectively. Each HLS directive is represented using 32 bits, wherein each directive option (unrolling factors, array partitions, etc.) is represented using 4 bits. The options are combined and zero-padded, if needed, to form the 32-bit representation for the directive.
Next, to prepare data for a network input, the binary values of all variables and directives in a solution are concatenated row-wise into a 2-D matrix. Advantages of using a fixed-point binary representation include seamlessly handling both continuous and discrete data types and achieving high precision while maintaining a dynamic range, thanks to the separation of integer and fractional values \cite{kalamkar2019BFLOAT16}. 

\begin{table*}[h]
    \caption{Example input configuration with 20 variables from HLSDataset. The variables are generated from the HLS report and synthesis results and include statistics like the number of digital signal processors (DSPs), block RAMs (BRAM), lookup tables (LUTs), flip-flops (FF), dynamic power (DP), shift register LUTs (SRL), etc.}
    \label{tab:Input20variables}
    \scriptsize
    \centering
    \begin{tabular}{|c|c|c|c|c|c|c|}
    \hline
    \textbf{Project} & \textbf{Clk-estimated(ns)} & \textbf{BRAM} & \textbf{DSP}& \textbf{FF}& \textbf{LUT}& \textbf{c-num-arith}\\ 
    \hline
    io1-l2n1n1-l4n1n1 & 8.419 & 32 & 5 & 727 & 1231 & 22\\ 
    \hline
    \textbf{c-num-logic}&\textbf{rtl-num-arith}& \textbf{rtl-num-logic}& \textbf{input-port} & \textbf{output-port} & \textbf{DP(mW)} & \textbf{Total LUTs}\\
    \hline
    10 & 19 &10 & 128 & 32 & 17.103 & 654 \\
    \hline
    \textbf{Logic LUTs}&\textbf{LUTRAMs}& \textbf{SRLs}& \textbf{FFs}& \textbf{RAMB36}&\textbf{RAMB18}&\textbf{DSP48}\\ 
    \hline
    654& 0 & 0 & 498 & 16 & 0 & 5 \\ 
    \hline
    \end{tabular}
\end{table*}

\subsection{Generative Models}\label{Sec:Generative models}
The second stage of Vaegan employs a generative model to create synthetic data. We explored a Variational Autoencoder (VAE) and a Generative Adversarial Network (GAN), but primarily use the VAE as it proved to be more effective for our purposes. We note that other state-of-the-art models like diffusion models can be used in this stage, and we plan to explore additional models in future work.

\subsubsection{Variational Autoencoder (VAE)}
A VAE \cite{kingma2022autoencoding} is a probabilistic directed graphical model defined by a joint distribution over a set of latent random variables $z$ and observed variables $x$, expressed as $p(x,z) = p(x|z)p(z)$. Fig. \ref{fig:methodology}(B) shows a sample VAE. In VAE, an \textit{encoder network} is leveraged to map the input variables to a continuous latent space. The parameters of a variational distribution defined within this latent space can spawn multiple different samples sharing the same underlying distribution. Typically, the prior distribution over the latent random variables, $p(z)$, is chosen as a standard Gaussian distribution, and the data likelihood $p(x|z)$ is generally a Gaussian or Bernoulli distribution whose parameters depend on $z$ through a deep neural network known as the \textit{decoder network}. This decoder is then employed to map the latent space back to the input space, leading to the generation of data points. The loss function ($L(x)$) of the encoder and decoder networks are jointly trained to maximize the evidence lower bound (ELBO):
\begin{align*}
L(x) = \mathbb{E}_{x\sim q(z|x)}[log(p(x|z))] - D_{KL}(q(z|x)\|p(z))
\end{align*}\label{eqa:VAE_LOSS}
where $D_{KL}$ is the Kullback–Leibler (KL) divergence between the latent distribution and standard normal distribution.

We employ a Multilayer Perceptron (MLP) network for both the encoder and decoder (MLPVAE). MLPVAE has two hidden layers and uses a ReLU activation function to provide non-linearity and symmetry for the VAE. A sigmoid activation function is used to reconstruct the input format for the decoder. MLPVAE encodes the real HLS data to a latent distribution by the mean and standard deviation and then decodes a sample from this distribution to reconstruct a new synthetic HLS data. We trained the model using binary cross entropy loss and Adam optimizer with an initial learning rate $\alpha = 0.0001$. A unique feature of the fixed-point input format used here is that we modified the loss function to place additional weight on changes to the most significant bit (MSB). This modification is made to account for the fact that each modification of the MSB causes a greater change in value compared to the lower bits.

\subsubsection{Generative Adversarial Network (GAN)}
A GAN \cite{goodfellow2020GAN}, exemplified in Fig. \ref{fig:methodology}(B), comprises a generative model trained through a competitive interplay between a generator and a discriminator network. The generator function $G(z)$ is typically initialized by drawing the latent variable $z$ from a basic prior distribution such as Gaussian, $p(z)$. The discriminator network, $D(x)$, outputs the probability of a given sample originating from the actual data distribution. It aims to differentiate between samples generated by the GAN and actual data. Concurrently, the generator endeavors to create samples with a high degree of realism, intending to deceive the discriminator into accepting these generated outputs as genuine. This iterative process between the generator and discriminator results in a zero-sum game, fostering an environment conducive to unsupervised learning. This contest between the two networks translates into a minimax problem where both networks strive to optimize their performances:

\begin{align*}
&\min_{G}\max_{D}\mathbb{E}_{x\sim p_{\text{data}}(x)}[\log{D(x)}] \\&+  \mathbb{E}_{z\sim p_{\text{z}}(z)}[1 - \log{D(G(z))}]
\end{align*}\label{eqa:GAN_LOSS}

We employed Deep Convolutional Generative Adversarial Network (DCGAN) \cite{radford2016unsupervised} which generally demonstrates superior performance over GAN in generating synthetic data. DCGAN uses convolutional neural networks (CNN) for both the generator and discriminator. The generator utilizes four deconvolution layers with Batch normalization and ReLU activation for the first three layers. A sigmoid activation function is used for the last layer to map to the [0,1] range for binary inputs. The generator takes a latent vector $z$ of size 100 from normal distribution as input and generates output matching the real data. The discriminator has four convolution layers with Batch normalization and LeakyReLU activation for the first three layers. Because the discriminator needs to output the probability of whether the input data is real or fake, a sigmoid activation function is also used in the last layer. Like MLPVAE, the DCGAN uses the same loss function as MLPVAE and Adam optimizer. We use the initial learning rate $\alpha = 0.07$ for the generator and $\alpha = 0.0001$ for the discriminator.

\subsection{Model Evaluation}
Fig. \ref{fig:methodology}(C) illustrates the evaluation process to verify the model's performance and the fidelity of the generated synthetic data to real data. The generated synthetic HLS data $\hat{x}$ is assessed relative to the real input data $x$ using a variety of state-of-the-art metrics to ensure a robust evaluation. These metrics include the Maximum Mean Discrepancy (MMD) \cite{gretton2012kernel}, Sum of the Square of the Distances (SSD), Percentage Root-mean-square Difference (PRD), and Cosine Similarity (COSS) \cite{li2023descod,3metrics}. We found the Fréchet Inception Distance (FID) \cite{heusel2017gans}, which is commonly used for evaluating GANs with 3-channel image inputs, to be incompatible with the 1-channel binary data format used in our work. Except for COSS, where higher values mean greater fidelity, smaller values in all other metrics indicate that the generated synthetic data closely resembles the real data.

\section{Experiments} \label{sec:experiments}
We performed experiments with input datasets from HLSDataset \cite{wei2023hlsdataset} as the original HLS data and used the MLPVAE and DCGAN models (Section \ref{Sec:Generative models}) to generate synthetic HLS data. We compared the Vaegan-generated data to the original data and synthetic data generated using two prior works. The first prior work \cite{LiaoDSE2023} (called `Gaussian' herein) randomly generated synthetic HLS data separately from all samples in each benchmark and variable based on a normal distribution. The second \cite{wharrie2022hapnest} (called `ABC' herein) generated genetic data using an approximate Bayesian computation (ABC) procedure. While this work did not generate HLS data, we found the approach used instructive for evaluating our work. We performed the experiments on an Intel i7 11700k @3.6GHz CPU with an NVIDIA RTX 3080 Ti GPU.

\subsection{Input}
We selected the HLSDataset due to its inclusion of partial post-implementation data, offering more diverse input data than db4hls \cite{ferretti2021db4hls}. HLSDataset targets two FPGA parts: xc7v585tffg1157-3 and xczu9eg-ffvb1156-2-i. We leveraged both parts to evaluate our approach's ability to generate distinct synthetic data for specific parts. HLSDataset provides several hardware configurations comprising HLS directives, HLS report estimation, post-synthesis, and partial post-implementation data. We separated the data into two categories: with and without HLS directives. This separation was necessary because each benchmark exhibited a different total number of loops and arrays, resulting in a varying number of HLS directives. Initially, we combined the data from the HLS report, post-synthesis, and post-implementation that did not include HLS directives. Of the 23 variables available per configuration, we selected 20 variables and eliminated 3 variables---target clock period of $10 ns$, clock uncertainty of $1.25 ns$, and frequency of $100 MHz$---that were fixed across all samples (see Table \ref{tab:Input20variables} for the details of variables). For the data with HLS directives, we used three directive options (loop pipeline, loop unrolling, and array partition) and eliminated the two (interface and resource) that were fixed for all benchmarks.

We collected a total of 9557 samples without HLS directives from two FPGA parts and 3717 samples with HLS directives for FPGA part xc7v585tffg1157-3 for Polybench. Of the 9 Polybench benchmarks, only 7 have directive files available and the directives are the same for both FPGA parts, hence the smaller number of samples. After selecting the variables from the dataset, we used a Python script\footnote{\url{https://github.com/yuchaoliao/VAEGAN/blob/main/DataTransformation.py}} to convert each variable to the 32-bit input format. 

\begin{table}[t]
    \caption{Hyperparameters for MLPVAE and DCGAN training}
    \label{tab:hyperparameters}
    \scriptsize
    \centering
    \begin{tabular}{ccccc}
    \hline
    \textbf{Model} & \textbf{MLPVAE} & \textbf{DCGAN-Generator} & \textbf{DCGAN-Descriminator}  \\ 
    \hline
    Learning Rate & 1e-4 & 7e-2 & 1e-4 \\ 
    
    Feature Map & 16 & 32 & 32 \\ 
    
    Batch Size & 20 & 20 & 20 \\ 
    
    Number of Epochs & 150 & 150 & 150 \\ 
    
    Optimizer  & Adam & Adam & Adam \\ 
    
    Layer Type & Linear & CNN & CNN \\
    
    Number of Layers  & 4 & 4 & 4 \\
    \hline
    \end{tabular}
\end{table}

\begin{table}[h]
    \caption{Evaluation of DCGAN and MLPVAE using MMD, SSD, PRD, and COSS for two FPGA parts without HLS directives. Larger numbers are better for COSS, while smaller are better for all other metrics. Results depict the mean over 5 runs of each model and the standard deviation ($\pm$).}
    \label{tab:numbermetrics}
    \scriptsize
    \centering
    \begin{tabular}{ccccc}
    \hline
    \textbf{Model} & \textbf{MMD} & \textbf{SSD} & \textbf{PRD\%}& \textbf{COSS}\\ 
    \hline
    \multicolumn{5}{c}{\textbf{FPGA part: xc7v585tffg1157-3}}\\  
    \hline
    Gaussian \cite{LiaoDSE2023} & $0.666\pm0.001$ & $99.697\pm0.087$ & $116.752\pm0.102$ & $0.390\pm0.001$\\ 
    
    ABC \cite{wharrie2022hapnest} & $0.667\pm0.002$ & $99.714\pm0.082$ & $116.754\pm0.033$ & $0.390\pm0.001$\\ 
    
    DCGAN & $0.614\pm0.002$ & $86.623\pm0.098$ & $110.707\pm0.055$ & $0.458\pm0.001$\\ 
    
    \textbf{MLPVAE} & $\textbf{0.370}\pm\textbf{0.002}$ & $\textbf{47.721}\pm\textbf{0.185}$ & $\textbf{82.830}\pm\textbf{0.302}$ & $\textbf{0.698}\pm\textbf{0.002}$\\ 
    \hline
    \multicolumn{5}{c}{\textbf{FPGA part: xczu9eg-ffvb1156-2-i}}\\
    \hline
    Gaussian \cite{LiaoDSE2023} & $0.662\pm0.001$ & $98.405\pm0.081$& $116.320\pm0.155$ & $0.393\pm0.001$\\ 
    
    ABC \cite{wharrie2022hapnest} & $0.663\pm0.003$ & $98.420\pm0.177$& $116.207\pm0.106$ & $0.394\pm0.001$\\
    
    DCGAN & $0.660\pm0.001$ & $87.215\pm0.100$ & $114.031\pm0.095$ & $0.436\pm0.001$\\ 
    
    \textbf{MLPVAE} & $\textbf{0.374}\pm\textbf{0.004}$ & $\textbf{47.641}\pm\textbf{0.166}$ & $\textbf{82.953}\pm\textbf{0.252}$ & $\textbf{0.696}\pm\textbf{0.001}$\\
    \hline
    \end{tabular}
\end{table}

\subsection{Model Configuration \& Training Parameters}

Table \ref{tab:hyperparameters} presents the hyperparameters used for the MLPVAE model, as well as for the generator and discriminator components of the DCGAN model. Without HLS directives, both models were trained separately on two FPGA parts, and synthetic HLS data was generated to match the real data size of 20 variables, each with 32 bits (20x32). Results (Section \ref{sec:results}) revealed the superiority of MLPVAE over both DCGAN and prior works. Thus, with HLS directives, for brevity, we report results for MLPVAE for part xc7v585tffg1157-3, with input widths of the number of directives (ranging from 19 to 35) plus 20 variables. The initial learning rate was the same $\alpha = 0.0001$ for both the generator and discriminator in DCGAN. We slowly increased the learning rate of the generator \cite{heusel2017gans} to $\alpha = 0.07$ until we found the balance between the generator and the discriminator. In each training epoch, we evaluated the model and generated synthetic HLS data using the MMD, SSD, PRD, and COSS metrics.

\begin{figure}[t]
\begin{subfigure}{1\linewidth}
  \centering
  \includegraphics[width=.8\textwidth]{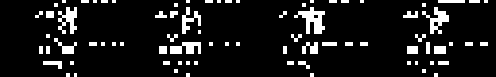}
  \caption{real data}
  \label{fig:visual_result_real}
\end{subfigure}
\begin{subfigure}{1\linewidth}
  \centering
  \includegraphics[width=.8\textwidth]{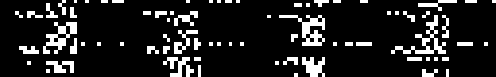} 
  \caption{Gaussian}
  \label{fig:visual_result_Random}
\end{subfigure}
\begin{subfigure}{1\linewidth}
  \centering
  \includegraphics[width=.8\textwidth]{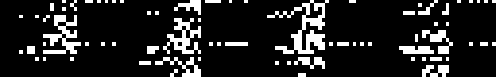} 
  \caption{ABC}
  \label{fig:visual_result_ABC}
\end{subfigure}
\begin{subfigure}{1\linewidth}
  \centering
  \includegraphics[width=.8\textwidth]{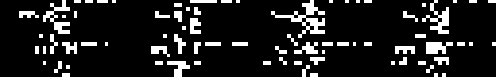} 

  \caption{DCGAN}
  \label{fig:visual_result_GAN}
\end{subfigure}
\begin{subfigure}{1\linewidth}
  \centering
  \includegraphics[width=.8\textwidth]{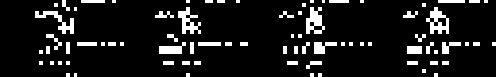} 
  \caption{MLPVAE}
  \label{fig:visual_result_VAE}
\end{subfigure}
\caption{Visualized HLS data comparison between (a) real data, (b) Gaussian \cite{LiaoDSE2023}, (c) ABC \cite{wharrie2022hapnest}, (d) DCGAN, and (e) MLPVAE for part xc7v585tffg1157-3 without HLS directives}
\label{fig:visual_result}
\end{figure}

\begin{table}[h]
    \caption{Evaluation of MLPVAE using MMD, SSD, PRD, and COSS for part xc7v585tffg1157-3 with HLS directives for seven benchmarks.}
    \label{tab:numbermetrics_with_HLS_directives}
    \scriptsize
    \centering
    \scalebox{1}{
    \begin{tabular}{ccccc}
    \hline
     \textbf{Benchamrk} & \multirow{2}{*}{\textbf{MMD}} & \multirow{2}{*}{\textbf{SSD}}  & \multirow{2}{*}{\textbf{PRD\%}} & \multirow{2}{*}{\textbf{COSS}}\\ 
     \textbf{(Directives)} &  & & & \\ 
    \hline
    \textbf{syrk(23)} & 
    $0.528\pm0.004$ & $57.009\pm0.193$ & $74.764\pm0.147$ & $0.745\pm0.001$\\ 
    
    \textbf{syr2k(36)} & 
    $0.489\pm0.006$ & $61.353\pm0.213$ & $70.074\pm0.163$ & $0.777\pm0.001$\\ 
    
    \textbf{k3mm(35)} & 
    $0.532\pm0.005$ & $58.597\pm0.315$ & $69.191\pm0.191$ & $0.781\pm0.001$\\ 
    
    \textbf{k2mm(35)} & 
    $0.516\pm0.004$ & $60.837\pm0.296$ & $70.027\pm0.186$ & $0.776\pm0.001$\\ 
    
    \textbf{gesummv(24)} & 
    $0.535\pm0.007$ & $52.539\pm0.282$ & $71.623\pm0.252$ & $0.765\pm0.001$\\ 
    
    \textbf{gemm(24)} & 
    $0.528\pm0.011$ & $57.643\pm0.304$ & $73.201\pm0.186$ & $0.757\pm0.001$\\ 
    
    \textbf{bicg(19)} & 
    $0.579\pm0.008$ & $55.230\pm0.219$ & $78.418\pm0.254$ & $0.723\pm0.001$\\ 
    
    \textbf{Average} & 
    \textbf{0.529} & \textbf{57.601} & \textbf{72.471} & \textbf{0.761}\\ 
    \hline
    \end{tabular}
    }
\end{table}

\subsection{Results} \label{sec:results}
Fig. \ref{fig:visual_result} depicts a visualization of the real HLS data without directives (for part xc7v585tffg1157-3) compared to the synthetic HLS data generated by our DCGAN and MLPVAE models and prior works (\textit{Gaussian} and \textit{ABC}).
As can be inferred from the visualization, both GAN and VAE are superior to the prior works in generating realistic synthetic data. A similar visualization with HLS directives (the figure is omitted for brevity) showed Vaegan's ability to generate data resembling the real data. However, we conducted additional quantitative assessments to more accurately evaluate the models.

Table \ref{tab:numbermetrics} provides a detailed evaluation of MLPVAE, DCGAN, and prior works (Gaussian and ABC) using MMD, SSD, PRD, and COSS metrics for data without HLS directives across both FPGA parts. MLPVAE consistently outperforms both DCGAN and prior methods across all metrics. The two prior works (Gaussian and ABC) generally perform similarly across all metrics. Specifically, regarding MMD, which represents the distribution score between real and synthetic data, MLPVAE generates synthetic HLS data that is on average 44.0\%, 44.1\%, and 41.6\% superior to that of Gaussian, ABC, and DCGAN for both FPGA parts. Considering the SSD, PRD, and COSS metrics, MLPVAE outperforms Gaussian by an average of 51.9\%, 28.9\%, and 78\%, respectively, and outperforms DCGAN by an average of 45.1\%, 26.2\%, and 56\%, respectively. 

DCGAN underperforms MLPVAE primarily due to the inherent challenges of fine-tuning. This stems from the sensitivity of DCGAN's convolutional layers to architectural choices (number of layers, filter sizes, strides), requiring extensive experimentation for optimal results. As a result, finding the right set of hyperparameters that leads to convergence can be less intuitive than with MLPVAE. To achieve high-fidelity synthetic HLS data, we fine-tuned our DCGAN's convolutional layers and hyperparameters significantly more often than our VAE's---over a hundred iterations for DCGAN compared to less than twenty for VAE, to achieve high-fidelity synthetic HLS data. On average, each experiment using DCGAN ran for 31 minutes and 46 seconds, whereas MLPVAE experiments ran for 11 minutes and 25 seconds. As such, compared to DCGAN, MLPVAE achieves better results while taking less time.

Table \ref{tab:numbermetrics_with_HLS_directives} shows the evaluation of MLPVAE on HLS data with HLS directives, using four metrics for the part xc7v585tffg1157-3 across seven benchmarks in Polybench. For brevity, we focus on MLPVAE due to its superiority over DCGAN and prior works.
We found that MLPVAE not only reduces training time (by up to 48$\times$) compared to DCGAN but also significantly improves the quality of results. MLPVAE generates synthetic data that resembles the real data, achieving average MMD, SSD, PRD, and COSS scores of 0.529, 57.601, 72.471, and 0.761, respectively.
These results show the promise of the proposed approach for generating high-fidelity synthetic data for HLS DSE. Furthermore, we observed that the results were benchmark-dependent because of variabilities in each benchmark's number of loops and loop levels, necessitating careful model tuning for each benchmark to optimize the synthetic data generation process.

\begin{figure}[t]
\centering
\includegraphics[width=0.90\columnwidth,keepaspectratio]{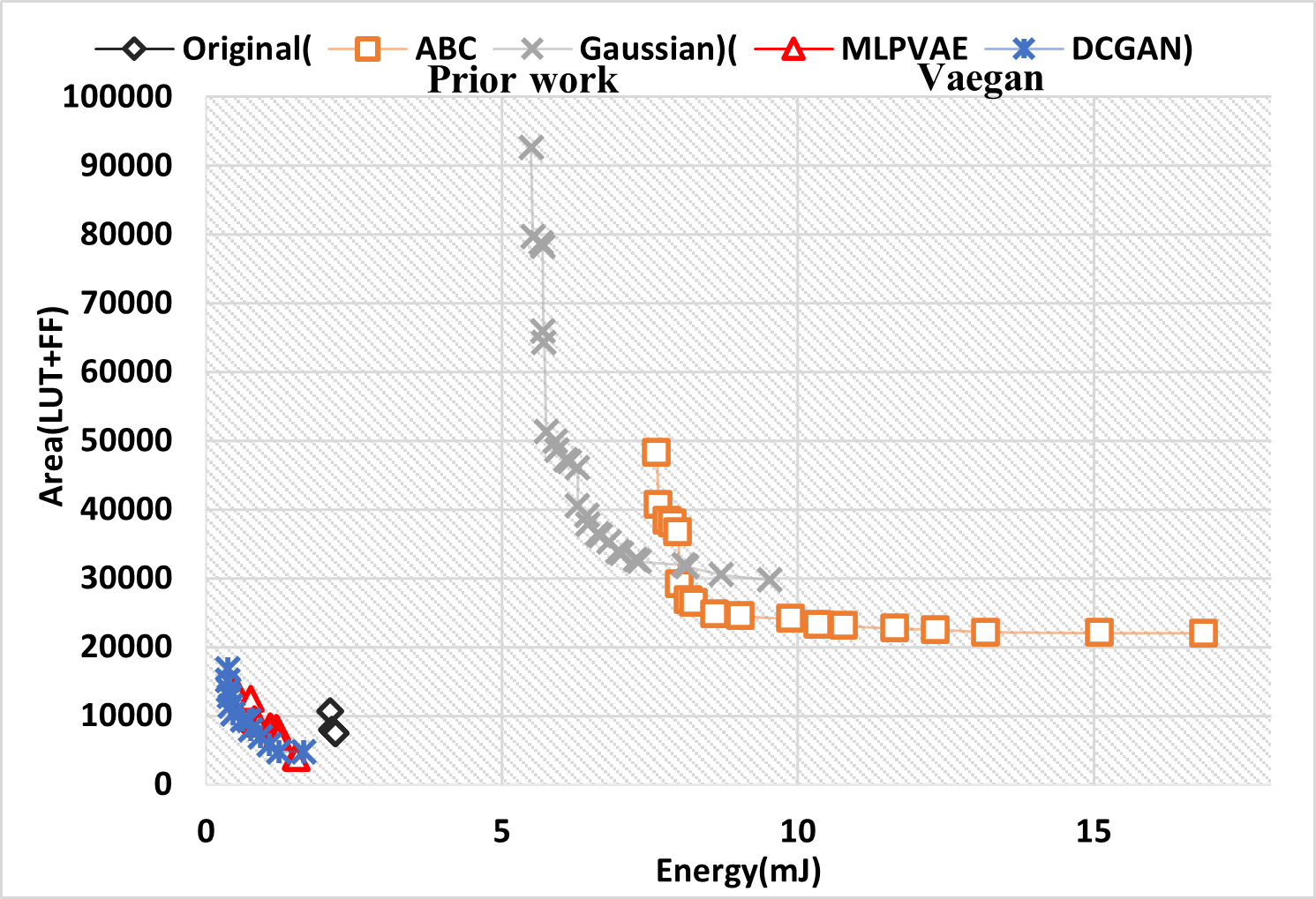}
\caption{Pareto-optimal design points for the area (FF+LUT) and energy of the original three-component wearable pregnancy monitoring system compared with prior work (ABC and Gaussian) and Vaegan (MLVPAE and DCGAN).}
\label{fig:Casestudy}
\end{figure}

\section{Case Study: Wearable Systems}
We briefly demonstrate and compare the use of Vaegan to generate synthetic data for a complex wearable pregnancy monitoring (WPM) device using real data with prior approaches. This case study involves a system-level DSE using a genetic algorithm to determine the Pareto-optimal energy and area under a constraint of the number of frequencies available on the target FPGA board. The WPM device comprises three components to monitor and process data for \textit{maternal heart rate}, \textit{blood oxygen saturation}, and \textit{abdomen contraction electromyography}. Each component runs with its period constraint, and communication is required between each component's controller and computation for each algorithm. The system-level design space comprises a combination of component design alternatives and their latencies, HLS directives to generate these design alternatives, and the interactions between the different components, resulting in a complex system-level DSE challenge. For the input HLS data, we used 69 manually generated post-implementation data points for the three-component WPM device. The design points are from different design alternatives (e.g., clock frequencies, latency constraints) yielding 1.92e+10 solutions, evaluated with four metrics: execution cycles, area (FF+LUT), critical path, and power. 

Due to space constraints, we omit the detailed description of the genetic algorithm, but it is modeled after a recent HLS DSE algorithm in \cite{LiaoDSE2023}. We use Vaegan, ABC, and Gaussian to generate four synthetic systems with the same size for the design alternatives. Fig. \ref{fig:Casestudy} presents the system-level Pareto-optimal configurations after applying the genetic algorithm to the original and synthetic systems using MLPVAE, DCGAN, ABC, and Gaussian. As seen in the figure, Vaegan generates a Pareto-optimal frontier that is much closer to the original system than ABC and Gaussian. We quantified the quality of the generated solutions using the Average Distance to Reference Set (ADRS) \cite{HLSsurveySchafer2022}. Compared to the original system, MLPVAE, DCGAN, ABC, and Gaussian achieved ADRS scores of 122.6\%, 124.1\%, 950.8\%, and 1353.6\%, respectively. A higher ADRS score implies a greater disparity from the original system in the Pareto frontier. MLPVAE modestly improved the ADRS over DCGAN by 1.23\%, and significantly outperformed ABC and Gaussian by 87.1\% and 90.9\%, respectively.

\section{Conclusion}
Existing benchmarks and datasets cannot cover all experimental conditions for system-level High-Level Synthesis (HLS) design space exploration (DSE). Consequently, utilizing synthetic data has emerged as a solution to this challenge. This paper proposes a novel approach---Vaegan---for generating realistic synthetic HLS data using generative machine learning models. To this end, we explored two kinds of models---a Multilayer Perception Variational Autoencoder (MLPVAE) and Deep Convolutional Generative Adversarial Networks (DCGAN)---and adapted them to generate synthetic HLS data. Experimental results show that Vaegan generates synthetic HLS data that closely mimics the ground truth distribution. Compared to prior work, Vaegan produced a Pareto front with superior design points, achieving energy and area values significantly closer to the original dataset. This demonstrates the effectiveness of our approach for improved design space exploration in complex systems.

Future work will explore the use of Vaegan as a machine learning predictor to bypass the High-Level Synthesis (HLS) process. Additionally, we will explore the integration of models such as the diffusion model into Vaegan for generating synthetic HLS data. This work will also be extended to accommodate various inputs, such as data flow graphs, for early-stage HLS exploration.

\begin{acks}
This work was partially supported by the Technology and Research Initiative Fund (TRIF) provided to the University of Arizona by the Arizona Board of Regents (ABOR) and National Science Foundation (NSF) Grant CNS-1844952. The work of R. Tandon was supported by NSF grants CAREER 1651492, CCF-2100013, CNS-2209951, and CNS-2317192.
\end{acks}

\balance
\bibliographystyle{ACM-Reference-Format}
\bibliography{References}

\end{document}